\documentclass[acmtog]{acmart}
\acmSubmissionID{1976}

\usepackage{booktabs} %

\usepackage[ruled]{algorithm2e} %
\usepackage{hyperref}       %
\usepackage{url}            %
\usepackage{booktabs}       %
\usepackage{amsfonts}       %
\usepackage{nicefrac}       %
\usepackage{microtype}      %
\usepackage{graphicx}
\usepackage{caption}
\usepackage{multirow}
\definecolor{bluegreen}{rgb}{0, 0.5, 0.5}
\usepackage{caption} %

\SetAlFnt{\small}
\SetAlCapFnt{\small}
\SetAlCapNameFnt{\small}
\SetAlCapHSkip{0pt}

\acmJournal{TOG}

\copyrightyear{2025}
\acmYear{2025}
\setcopyright{acmlicensed}\acmConference[SA Conference Papers '25]{SIGGRAPH Asia 2025 Conference Papers}{December 15--18, 2025}{Hong Kong, Hong Kong}
\acmBooktitle{SIGGRAPH Asia 2025 Conference Papers (SA Conference Papers '25), December 15--18, 2025, Hong Kong, Hong Kong}
\acmDOI{10.1145/3757377.3763956}
\acmISBN{979-8-4007-2137-3/2025/12}

\begin{document}
\title{DreamO: A Unified Framework for Image Customization}

\author{Chong Mou}
\orcid{0000-0003-0296-4893}
\affiliation{%
 \institution{Intelligent Creation Team, ByteDance; Peking University}
 \country{China}}
\email{eechongm@gmail.com}

\author{Yanze Wu}\authornote{Correspondence: Yanze Wu <wuyanze123@gmail.com>}
\orcid{0000-0001-6648-5387}\authornote{Project leads}
\affiliation{%
 \institution{Intelligent Creation Team, ByteDance}
 \country{China}}
\email{wuyanze123@gmail.com}

\author{Wenxu Wu}
\orcid{0009-0002-9622-3665}
\affiliation{%
 \institution{Intelligent Creation Team, ByteDance}
 \country{China}}

\author{Zinan Guo}
\orcid{0009-0002-7649-4097}
\affiliation{%
 \institution{Intelligent Creation Team, ByteDance}
 \country{China}}

\author{Pengze Zhang}
\orcid{0000-0002-3736-5295}
\affiliation{%
 \institution{Intelligent Creation Team, ByteDance}
 \country{China}}

 \author{Yufeng Cheng}
\orcid{0009-0001-0918-3603}
\affiliation{%
 \institution{Intelligent Creation Team, ByteDance}
 \country{China}}

 \author{Yiming Luo}
\orcid{0009-0007-8216-0348}
\affiliation{%
 \institution{Intelligent Creation Team, ByteDance}
 \country{China}}

 \author{Fei Ding}
\authornotemark[2]
\orcid{0009-0004-1880-9681}
\affiliation{%
 \institution{Intelligent Creation Team, ByteDance}
 \country{China}}

 \author{Shiwen Zhang}
\orcid{0009-0005-1580-3396}
\affiliation{%
 \institution{Intelligent Creation Team, ByteDance}
 \country{China}}

 \author{Xinghui Li}
\orcid{0009-0009-9813-6259}
\affiliation{%
 \institution{Intelligent Creation Team, ByteDance}
 \country{China}}

 \author{Mengtian Li}
\orcid{0000-0001-6724-6177}
\affiliation{%
 \institution{Intelligent Creation Team, ByteDance}
 \country{China}}

  \author{Mingcong Liu}
\orcid{0000-0003-1638-8348}
\affiliation{%
 \institution{Intelligent Creation Team, ByteDance}
 \country{China}}

  \author{Yunsheng Jiang}
\orcid{0009-0001-3128-7521}
\affiliation{%
 \institution{Intelligent Creation Team, ByteDance}
 \country{China}}

  \author{Shaojin Wu}
\orcid{0000-0002-7899-0863}
\affiliation{%
 \institution{Intelligent Creation Team, ByteDance}
 \country{China}}

  \author{Songtao Zhao}
\orcid{0009-0001-1399-0651}
\affiliation{%
 \institution{Intelligent Creation Team, ByteDance}
 \country{China}}

  \author{Jian Zhang}
\orcid{0000-0001-5486-3125}
\affiliation{%
 \institution{Peking University}
 \country{China}}
 
   \author{Qian He}
\orcid{0009-0000-9978-7904}
\affiliation{%
 \institution{Intelligent Creation Team, ByteDance}
 \country{China}}
 
   \author{Xinglong Wu}
\orcid{0009-0007-5860-5941}
\affiliation{%
 \institution{Intelligent Creation Team, ByteDance}
 \country{China}}

\begin{teaserfigure}
  \includegraphics[width=\textwidth]{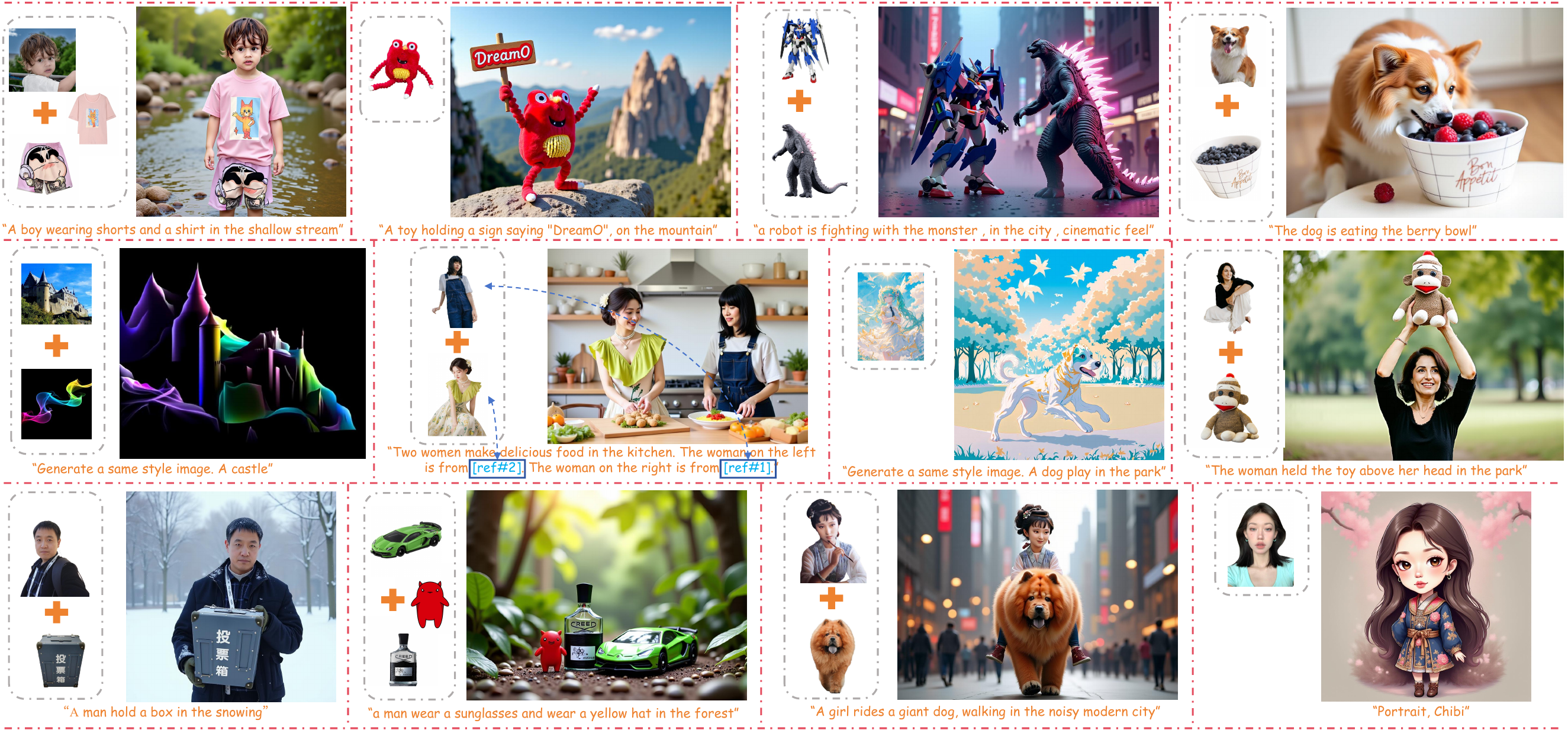}
  \caption{The image customization capability of our proposed DreamO.}
  \label{fig:teaser}
\end{teaserfigure}

\renewcommand\shortauthors{Mou et al}

\begin{abstract}
Recently, extensive research on image customization (e.g., identity, subject, style, background, etc.) demonstrates strong customization capabilities in large-scale generative models. However, most approaches are designed for specific tasks, restricting their generalizability to combine different types of condition. Developing a unified framework for image customization remains an open challenge. In this paper, we present \textbf{DreamO}, an image customization framework designed to support a wide range of tasks while facilitating seamless integration of multiple conditions. Specifically, DreamO utilizes a diffusion transformer (DiT) framework to uniformly process input of different types. During training, we introduce a feature routing constraint to facilitate the precise querying of relevant information from reference images. Additionally, we design a placeholder strategy that associates specific placeholders with conditions at particular positions, enabling control over the placement of conditions in the generated results. Moreover, we employ a progressive training strategy to ensure smooth model convergence and correct the generation quality of the final output. Extensive experiments demonstrate that the proposed DreamO can effectively perform various image customization tasks with high quality and flexibly integrate different types of control conditions.
Project page: \href{https://mc-e.github.io/project/DreamO}{https://mc-e.github.io/project/DreamO}
\end{abstract}



\maketitle

\section{Introduction}
Due to the high-quality image generation and stable performance of diffusion models~\cite{ddpm}, substantial research efforts focus on controllable generation by leveraging their generative priors~\cite{t2iadapter,controlnet}. Among these, image customization aims to ensure that generated outputs remain consistent with a reference image in specific attributes, such as identity~\cite{instantid,fastcomposer,pulid}, object appearance~\cite{realcustom,dreambooth,anystory}, virtual try-on~\cite{tryon-1,tryon-2,tryon-3}, and style~\cite{csgo,deadiff,stylealign}. Despite the abundance of task-specific approaches, developing a unified framework for image customization remains a challenge.

Early research Composer~\cite{composer} jointly trains a diffusion model with multi-condition input,~\textit{e.g.}, depth, color, sketch. Some methods~\cite{unicontrol-1, unicontrol-2} train additional control blocks~\cite{controlnet,t2iadapter} to support general spatial control on the generation result, which greatly saves training costs. However, their control ability is restricted to some simple spatial conditions, and the interactions between conditions are rigid and have control redundancy. Recently, the DiT~\cite{dit} framework greatly scales up the performance of diffusion models. Based on DiT, OminiControl~\cite{ominicontrol} proposes to integrate image conditions by unified sequence with diffusion latent. It can perform various customization tasks, \textit{e.g.}, identity, color, and layout. Despite its advantages, OminiControl is trained separately on different tasks, struggling to process multiple conditions. Recently, OmniGen~\cite{omnigen} trains a general generation control based on a pre-trained large language model~\cite{phi3} (LLMs). UniReal~\cite{unireal} achieves this through video generation pretraining followed by full-model post-training. However, we argue that high-quality, multi-concept image customization cannot be achieved by merely leveraging the general capabilities of large language models like OmniGen or video models such as UniReal. Instead, it requires specialized architectural designs. Currently, the research community lacks an efficient and effective method specifically tailored for image customization under multi-concept and multi-conditional scenarios.

In this paper, we design a unified image customization approach based on a pre-trained DiT. With a low training cost and a single model, our method can support various conditions (\textit{e.g.}, identity, subject, try-on, and style) and enables interactions among different kinds of condition, as shown in~\ref{fig:teaser}. Specifically, we follow the unified sequence conditioning format introduced in OminiControl~\cite{ominicontrol}, and introduce a routing constraint on the internal representations of DiT during training. This ensures content fidelity and promotes the disentanglement of different control conditions. A placeholder strategy is also designed to enable control over the placement of conditions in the generated results. In addition, we construct large-scale training data covering multiple tasks and design a progressive training strategy. This enables the model to progressively acquire robust and generalized image customization capabilities.

In summery, this paper has the following contributions:

\begin{itemize}
    \item We propose DreamO, a unified framework for image customization. It achieves various complex and multi-condition customization tasks by training a small set of additional parameters on a pre-trained DiT model.
    \item Based on representation correspondences within the diffusion model, we design a feature routing constraint to enhance consistency fidelity and enable effective decoupling in multi-condition scenarios.
    \item We introduce a progressive training strategy to facilitate convergence in multi-task and complex task settings. Moreover, we design a placeholder strategy to establish correspondence between textual descriptions and condition images.
    \item 
    Extensive experiments demonstrate that our method not only produces high-quality results in various image customization tasks, but also exhibits strong flexibility in multi-condition scenarios.
\end{itemize}

\begin{figure*}[t]
\centering
\begin{minipage}[t]{\linewidth}
\centering
\includegraphics[width=.9\columnwidth]{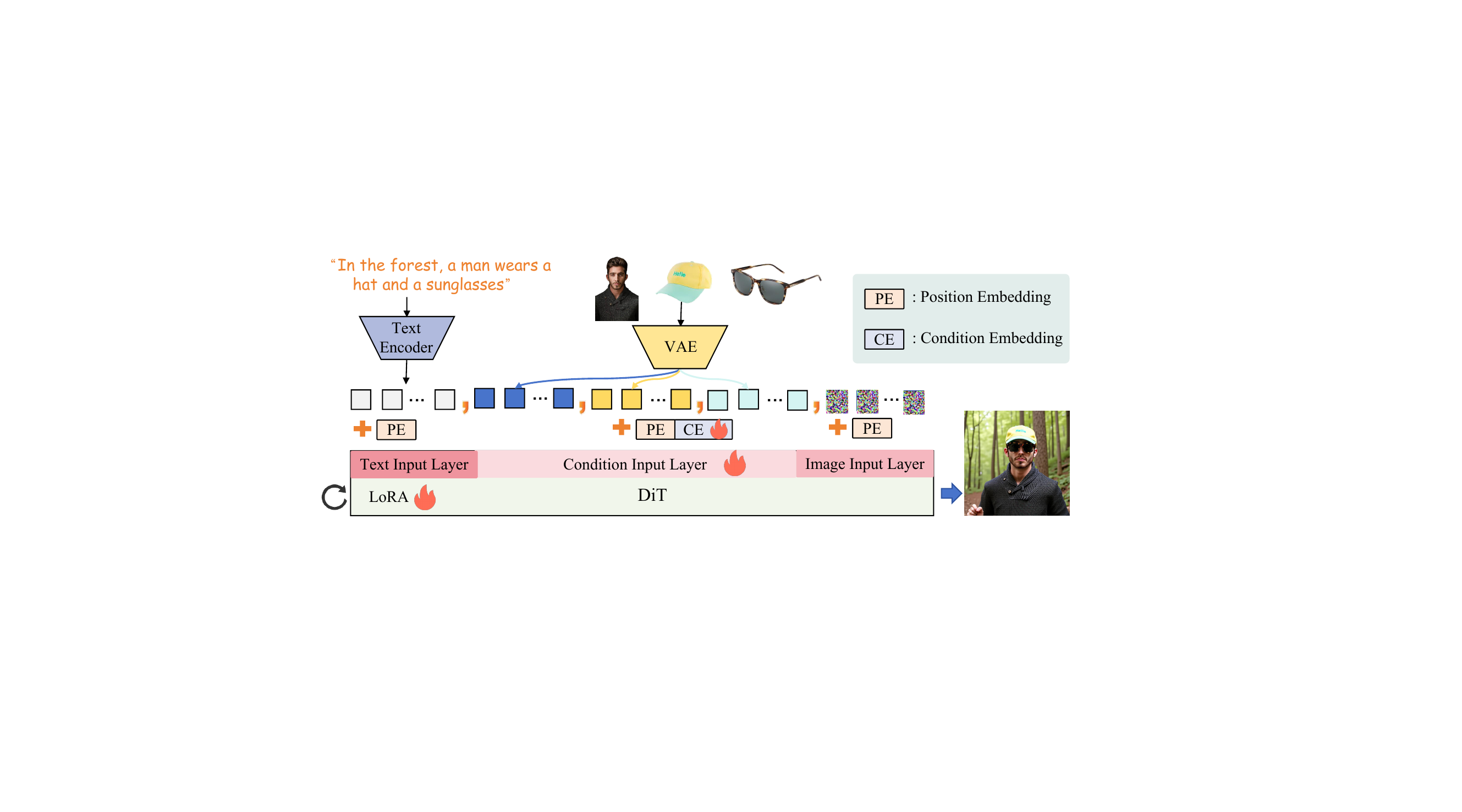}
\end{minipage}
\centering
\vspace{-16pt}
\caption{Overview of our proposed DreamO, which can uniformly handle commonly used consistency-aware generation control.
}
\vspace{-10pt}
\label{overview} 
\end{figure*}

\section{Related Works}
\subsection{Diffusion Models}
As a powerful generation paradigm, the diffusion model~\cite{ddpm,diff_beat} rapidly dominates the image generation community. Its high generation quality and stable performance have been successfully applied to various tasks \textit{, e.g.}, text-to-image generation~\cite{t2i1,dall-e2,glid,ldm} and image editing~\cite{p2p,sdedit,paint-by,dragondiffusion}. Some strategies, such as latent diffusion~\cite{ldm} and flow matching sampling~\cite{flow}, are proposed to enhance the performance. Most early works~\cite{ldm, glid} utilize the UNet architecture as the diffusion model. Recently, the diffusion transformer~\cite{dit} (DiT) architecture has emerged as a superior choice, offering improved performance through straightforward scalability. 

\subsection{Controllable Image Generation}
Advancements in diffusion models drive the rapid development of controllable image generation. In this community, text is the most fundamental conditioning input, \textit{e.g.}, Stable Diffusion~\cite{ldm}, Imagen~\cite{imagen}, and DALL-E2~\cite{dall-e2}. To achieve accurate spatial control, some methods, \textit{e.g.}, ControlNet~\cite{controlnet} and T2I-Adapter~\cite{t2iadapter}, propose adding control modules on pre-trained diffusion models. UniControl~\cite{unicontrol-1,unicontrol-2} propose to unify different spatial conditions with a joint condition input. In addition to spatial control, some unspatial conditions are also studied. The IP-Adapter~\cite{ipadapter} utilizes cross-attention to inject image prompt in the diffusion model to control some unspatial properties, \textit{e.g.}, identity and style. In addition to these representative works, other related attempts~\cite{relat_1,anystory,relat_3,relat_4,relat_5,relat_6,relat_7} also help to broaden the scope of controllable image generation. 

Recent advancements in DiT-based diffusion models further promote the development of controllable image generation. For instance, in-context LoRA~\cite{Incontext} and OminiControl~\cite{ominicontrol} introduce a novel approach by concatenating all input tokens (\textit{i.e.}, text, image, and conditions) and training LoRA with task-specific datasets for various applications. Subsequently, OmniGen~\cite{omnigen} and UniReal~\cite{unireal} optimize the entire diffusion model in multiple stages on larger-scale training data, achieving improved understanding of input conditions.

\subsection{Cross-attention Routing in Diffusion Models}
\label{cross_route}
Existing studies (\textit{e.g.}, Prompt-to-Prompt~\cite{p2p}) demonstrate that text-visual cross-attention maps inherently establish spatial-semantic correspondence between linguistic tokens and visual generation, \textit{i.e.}, the similarity response aligns with the spatial region of the corresponding subject in the generated result. 
Building upon this observation, UniPortrait~\cite{uniportrait} constrains the influence region of condition features for identity-specific generation in multi-face scenarios, and AnyStory~\cite{anystory} further extends this approach to subject-driven generation. Some recent works~\cite{seg_att} show that the cross-attention map in the DiT framework also exhibits spatial properties. In this paper, we explore routing constraints in the DiT framework.

\section{Method} 

\subsection{Preliminaries}
The Diffusion Transformer (DiT) model~\cite{dit} employs a transformer as the denoising network to refine diffusion latent. Specifically, in the input, the 2D image latent $\mathbf{z}_{t}\in\mathbb{R}^{c\times w\times h}$ is patchified into a sequence of 1D tokens $\mathbf{z}_{t}\in\mathbb{R}^{c\times (\frac{w}{p}\times \frac{h}{p})}$, where $(w, h)$ is spatial size, $c$ is the number of channels, and $p$ is the patch size. The image and text tokens are concatenated and processed by the DiT model in a unified manner. Apart from model architectures, more efficient sampling strategies (\textit{e.g.}, Flow Matching~\cite{flow}) are also proposed. Unlike DDPM~\cite{ddpm}, Flow Matching conducts the forward process by linearly interpolating between noise and data in a straight line. At the time step $t$, latent $\mathbf{z}_t$ is defined as: $\mathbf{z}_t = (1-t)\mathbf{z}_0+t\epsilon$, where $\mathbf{z}_0$ is the clean image, and $\epsilon\in \mathcal{N}(0,1)$ is the Gaussian noise. The model is trained to directly regress the target velocity given the noised latent $\mathbf{z}_t$, timestep t, and condition $y$. The objective is to minimize the mean squared error:
\begin{equation}
\label{loss}
    L_{diff} = E[||(\mathbf{z}_0-\epsilon)-\mathcal{V}_{\theta}(\mathbf{z}_t, t, y)||^2],
\end{equation}
where $\mathcal{V}_{\theta}$ refers to the diffusion model. The DiT framework and Flow Matching are widely used in some recent diffusion models, such as Stable Diffusion3~\cite{sd3} and Flux~\cite{flux}.

\subsection{Overview}
An overview of our method is presented in Fig.~\ref{overview}. Specifically, we utilize the Flux-1.0-dev~\cite{flux} as the base model to build a unified framework for image customization, \textit{e.g.}, style, identity, subject appearance, and try-on. Given $n$ condition images $\mathbf{C}=\{\mathbf{C}_1, ...,\mathbf{C}_n\}$, we first reuse the VAE~\cite{vae} of Flux to encode the condition image to the same latent space as noisy latent. Note that the size of the condition image is flexible. Higher resolutions are recommended for detail-rich images to preserve clarity, while lower resolutions are sufficient for images with fewer details, thereby reducing compression costs. Then, all tokens (\textit{i.e.}, image, text, condition) are concatenated along the sequence dimension and fed into Flux. 
To enable the model to incorporate the condition input, we introduce a condition mapping layer at the input of Flux. 
The position embedding (PE) of condition tokens is aligned with that of the noisy latent using Rotary Position Embedding (RoPE)~\cite{rope}. Inspired by non-overlapping position embedding in OminiControl~\cite{ominicontrol}, we extend these embeddings along the diagonal in a similar fashion. In addition, we introduce a trainable and index-wise condition embedding (CE) $\mathbb{R}^{10\times c}$, which is directly added to condition tokens. 
Following OminiControl~\cite{ominicontrol}, we integrate Low-Rank Adaptation (LoRA)~\cite{lora} modules into Flux as trainable parameters.

\begin{figure}[t]
\centering
\begin{minipage}[t]{\linewidth}
\centering
\includegraphics[width=\columnwidth]{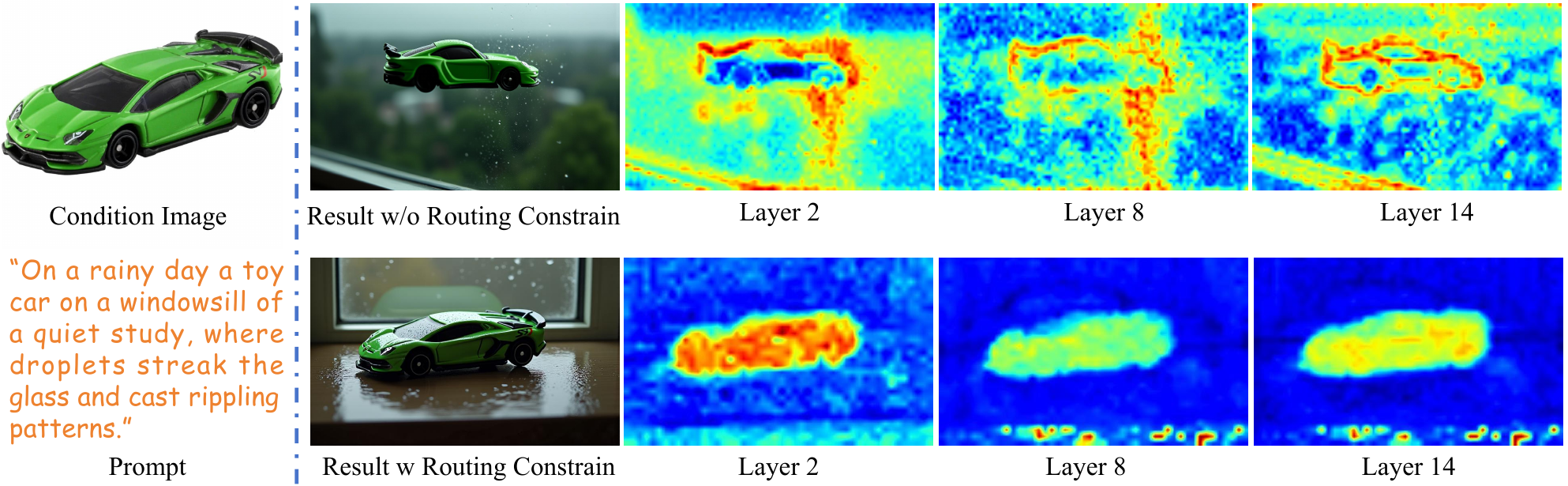}
\end{minipage}
\centering
\vspace{-10pt}
\caption{Visualization of cross-attention maps in subject-driven image generation. The first row shows results from a model trained without routing constraints, while the second row presents results from a model trained with routing constraints.
}
\label{attmap} 
\end{figure}

\subsection{Routing Constraint}
Inspired by UniPortrait~\cite{uniportrait} and AnyStory~\cite{anystory}, in this paper, we design routing constraint in the DiT framework for general image customization tasks. As illustrated in Fig.~\ref{overview}, within the condition-guided framework, cross-attention exists between condition images and the generation result:
\begin{equation}
    \mathbf{M}=\frac{\mathbf{Q}_{cond,i}\mathbf{K}_{img}^{T}}{\sqrt{d}},
\end{equation}
where $\mathbf{Q}_{cond,i}\in \mathbb{R}^{l_{cond,i}\times c} $ refers to the condition tokens of the $i$-$th$ condition image. $\mathbf{K}_{img}\in \mathbb{R}^{l\times c}$ is the tokens of noisy image latent. The cross-attention map $\mathbf{M}\in \mathbb{R}^{l_{cond,i}\times l}$ is a dense similarity between the $i$-$th$ condition image and the generation result. To obtain the global response of the condition image in different locations of the generated output, we average the dense similarity matrix along the $l_{cond,i}$ dimension, resulting in a response map $\mathbf{M}\in \mathbb{R}^{l}$, representing the global similarity of the condition image on the generated result. To constrain the image-to-image attention focus on the specific subject, MSE loss is employed to optimize the attention within DiT across condition images and the generation result:
\begin{equation}
\label{route_loss}
    L_{route} = \frac{1}{n_c\times n_l}\sum_{j=0}^{n_l-1}\sum_{i=0}^{n_c-1}||\mathbf{M}_{i}^j-\mathbf{M}_{target,i}||_2^2,
\end{equation}
where $i$ and $j$ refer to the condition index and layer index. $n_c$ and $n_l$ is the number of conditions and number of layers, respectively. $\mathbf{M}_{target}$ refers to the subject mask for the target image. As shown in the second row of Fig.~\ref{attmap}, after training with routing constraint, the attention of the condition image clearly focuses on the target subject, and the result shows improved consistency with the reference image in terms of details. In addition to improved consistency, this strategy also helps decoupling in multi-reference cases.

In addition to image-to-image routing constraint, we also design placeholder-to-image routing constraint to establish correspondences between textual descriptions and condition inputs. Specifically, for the $i$-$th$ condition, we append a placeholder $[ref\#i]$ after the corresponding instance name, \textit{e.g.}, \textcolor{orange}{``A\ women\ from}\ \textcolor{blue}{[ref\#1]}\ \textcolor{orange}{and\ a\ woman\ from}\ \textcolor{blue}{[ref\#2]}\ \textcolor{orange}{is\ walking\ in\ the\ park.''}. During the training for multi-condition tasks, we calculate the similarity between the conditional image tokens and the placeholder tokens. The routing constraint ensures that the similarity between $\mathbf{C}_i$ and $[ref\#i]$ is $1$, while it is $0$ for all other pairs:
\begin{equation}
\label{loss_placeholder}
    L_{holder}=\frac{1}{n_c}\sum_{i=0}^{n_c-1}||Softmax(\mathbf{Q}_{cond,i}\times \mathbf{K}_{text,i}^T)-\mathbf{B}_i||_2^2,
\end{equation}
where $\mathbf{K}_{text,i}$ refers to the text feature of $[ref\#i]$. $\mathbf{B}_i$ is a binary matrix, where the value is $1$ when the placeholder matches the condition image, and $0$ otherwise.

The final loss function of our method is defined as:
\begin{equation}
    L = \lambda_{diff}\cdot L_{diff}+\lambda_{route}\cdot L_{route}+\lambda_{holder}\cdot L_{holder},
\end{equation}
where $\lambda_{diff}$, $\lambda_{route}$ and $\lambda_{holder}$ are loss weights. To allow the model to handle regular text input, we introduce normal text without placeholders with a probability of $50\%$ and discard $L_{holder}$ accordingly. Note that $L_{route}$ and $L_{holder}$ do not incur significant additional computational overhead during training (2.5s/iter vs. 3s/iter).

\begin{figure}[t]
\centering
\begin{minipage}[t]{\linewidth}
\centering
\includegraphics[width=1\columnwidth]{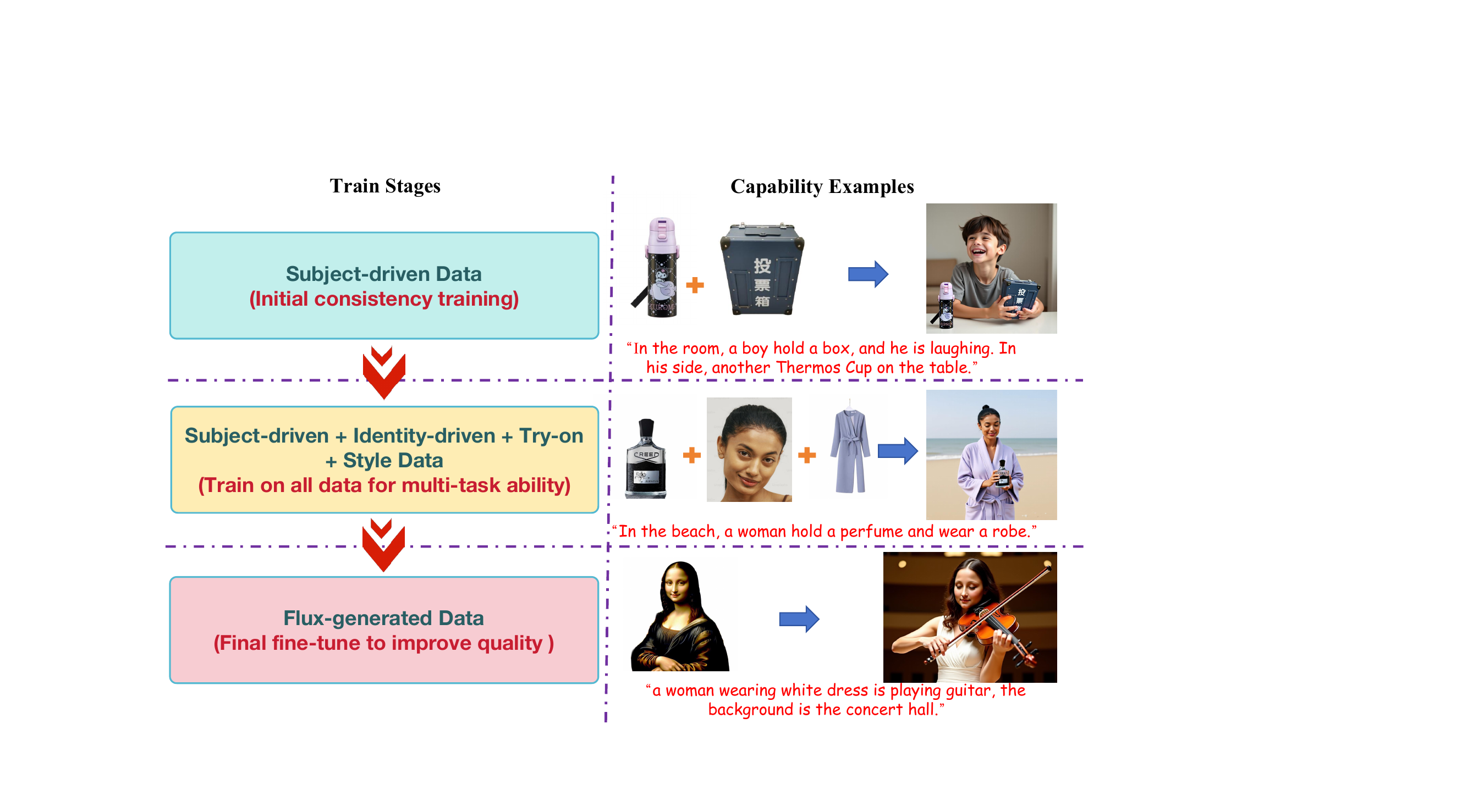}
\end{minipage}
\centering
\vspace{-18pt}
\caption{
The progressive training pipeline of our method. Left column shows the three training stages of our method. Right column shows the generation capability after the training of each stage. 
}
\label{train_stages} 
\end{figure}

\begin{figure*}[t]
\centering
\begin{minipage}[t]{\linewidth}
\centering
\includegraphics[width=1\columnwidth]{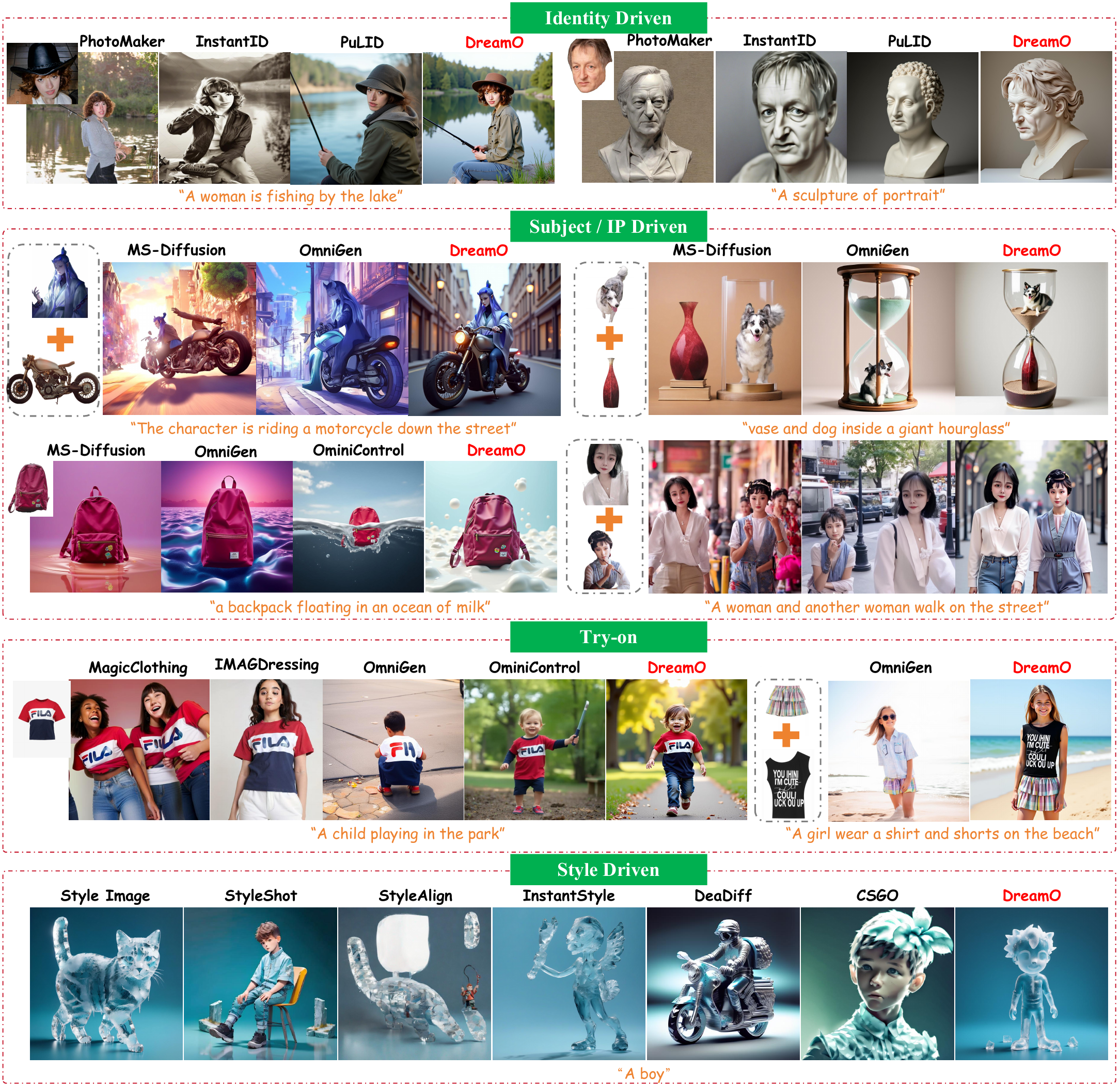}
\end{minipage}
\centering
\caption{
 Visual comparison between our DreamO and other methods.
}
\label{vis_cmp} 
\end{figure*}

\subsection{Training Data Construction}
\label{two-col}
To achieve generalized image customization, we collect training data, covering a wide range of tasks.

\noindent \textbf{Identity paired data}. Since high-quality identity paired data~\cite{photomaker} is difficult to collect from the Internet, we adopt the open-source ID customization method PuLID~\cite{pulid} for dataset construction. Specifically, we generate two images of the same identity using PuLID-FLUX, which then serve as mutual references. We also provide PuLID-SDXL with a reference face image and a text prompt describing the desired style to produce stylized training pairs. Finally, we collect 150K photorealistic data and 60K stylized identity data.

\noindent \textbf{Subject-driven data}. For single-subject-driven image customization, we utilize the Subject200K~\cite{ominicontrol} dataset as part of training data. To rectify the absence of character-related conditions, we collect 100K paired character-related data through retrieval. For multi-subject-driven image customization, we construct some two-column images through concatenation on the Subject200K dataset. 
In addition, we employ X2I-subject~\cite{omnigen} dataset in multi-subject-driven training. To improve human-driven generation, we develop a pipeline similar to MovieGen~\cite{moive_gen}. Starting with a long-video dataset, we apply content-aware scene detection to extract short clips. Mask2Former~\cite{mask2former} is used to generate human masks for key frames and perform object tracking. For cross-clip instance matching, we use SigLip~\cite{zhai2023sigmoid} embeddings and clustering. 

\noindent \textbf{Try-on data}. We create a paired try-on dataset from two sources. One part consists of paired model and clothing images collected directly from the Web. For the other, we first crawl high-quality model images as ground truth, then use image segmentation~\cite{jin2023sssegmenation,jin2024idrnet} to extract clothing items and generate the corresponding pairs. All images are manually filtered to remove low-quality samples, resulting in a dataset of 500K try-on pairs.

\noindent \textbf{Style-driven data}. This paper tackles two style transfer tasks: (1) style reference control with text descriptions of content, and (2) style reference with content reference images. For the first task, we use an internal style customization model based on SDXL to generate images with the same style but different content from two distinct prompts. For the second task, training requires style reference images, content reference images, and target images, where the target image must match both the style and content structure of the reference images. Based on the type 1 dataset, we produce the content reference for each style image using Canny-guided Flux~\cite{flux_canny}. More details are provided in the \textbf{Appendix}.

\noindent \textbf{Routing mask extraction}. To obtain the labels for the routing constraint (i.e., Eq.\ref{route_loss}), we use
LISA\cite{lisa} to extract mask of object conditioned on the text descriptions. In certain complex datasets, we employ InternVL~\cite{internvl} to generate description of the target object. More details are provided in the \textbf{Appendix}.

\begin{figure*}[t]
\centering
\begin{minipage}[t]{\linewidth}
\centering
\includegraphics[width=1\columnwidth]{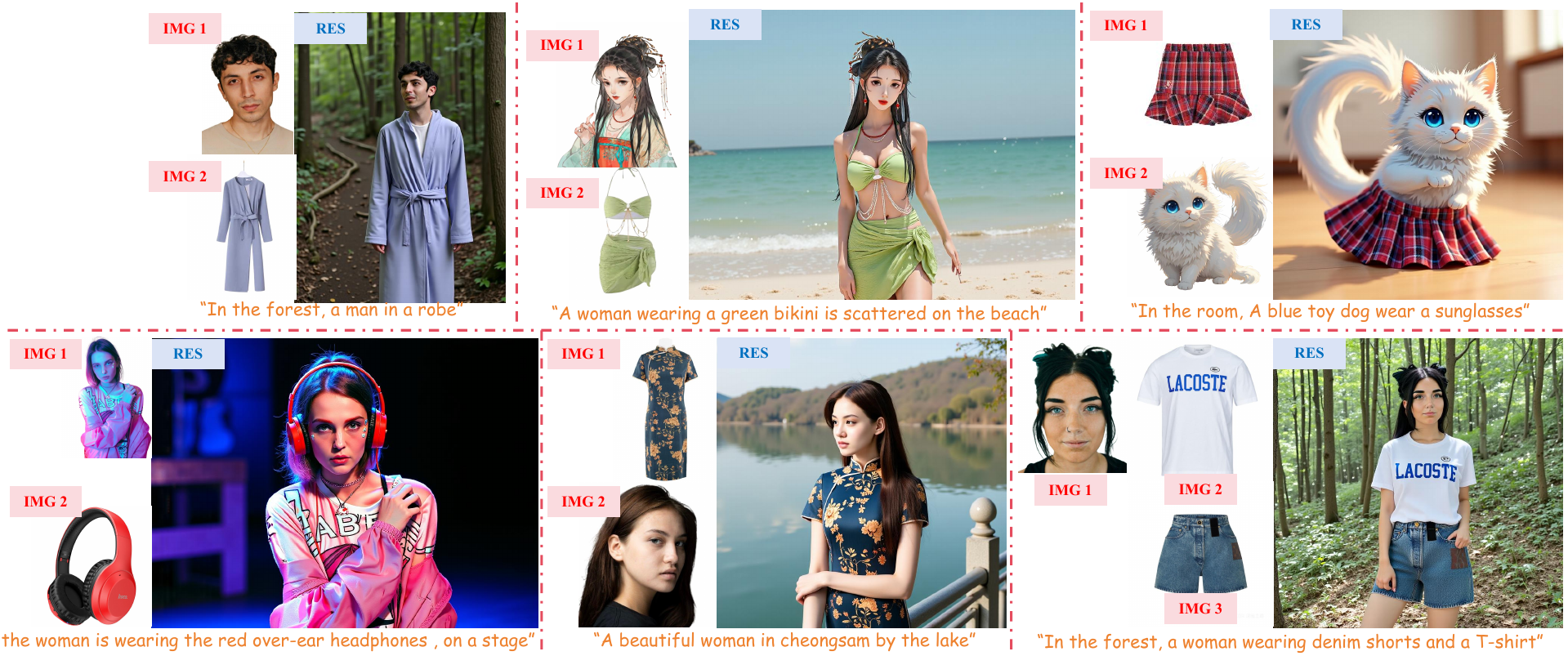}
\end{minipage}
\centering
\caption{
The capability of our proposed DreamO in try-on image customization.
}
\label{show_3} 
\end{figure*}

Although the training data is constructed separately for different tasks, we observe the emergence of some cross-task capabilities. For instance, the model can customize the combination of ID and Try-on (as shown in Fig.~\ref{show_3}), which does not exist in the training data.

\subsection{Progressive Training Process}
In experiments, we find that training directly on all data makes convergence difficult. This is mainly due to the limited capacity of LoRA~\cite{lora} optimization, making it difficult for the model to capture task-specific capabilities under complex data distributions. In addition to convergence, the quality of the output after training deviates from the original prior of Flux~\cite{flux}. This divergence is caused by the impact of some low-quality training samples.

\begin{table*}[t]
\captionsetup{type=table}
\caption{Quantitative evaluation of subject-driven customization.}
\centering
\begin{tabular}{c | c c c c | c c c}
\toprule
 & \multicolumn{4}{c|}{Single-subject Customization} & \multicolumn{3}{|c}{Multi-subject Customization}\\
 & MS-Diffusion & OmniGen & OminiControl & DreamO & MS-Diffusion & OmniGen & DreamO \\
\toprule
CLIP-sim $\uparrow$ & 0.8989 &  0.8824 & 0.8220 & \textbf{0.9150}& 0.7686 & 0.7605 & \textbf{0.7775} \\
DINO-sim $\uparrow$ & 0.7746 &  0.7582 & 0.6089 & \textbf{0.8056} & 0.6113 & 0.5646 & \textbf{0.6253}\\
Text-sim $\uparrow$ & 31.78 & 31.74 & 31.12 & \textbf{31.92} & 31.34 & 29.55 & \textbf{31.46}\\
\toprule
\end{tabular}
\label{tb_cp_subject}
\end{table*}

To address these issues, we design a progressive training strategy that allows the model to smoothly converge across different tasks while mitigating the influence of training data on the generation prior of Flux. The training pipeline is shown in Fig.~\ref{train_stages}. Specifically, we first optimize the model on subject-driven training data to initiate the model with a consistency-preserving capability. Note that the Subject200K~\cite{ominicontrol} training data is generated by the base model (\textit{i.e.}, Flux), thus shares a similar distribution with the model generation space, which facilitates fast convergence. Since the X2I-subject~\cite{omnigen} dataset is synthetically generated by MS-Diffusion~\cite{ms_diffusion}, a lot of training samples contain undesired artifacts and distortions. Therefore, during this warm-up stage, the two-column Subject200K images described in Sec.~\ref{two-col} are also utilized as part of the training data to facilitate rapid convergence of the multi-reference control. The right part of Fig.~\ref{train_stages} illustrates that after the first training stage, the model acquires an initial subject-driven generation capability and presents strong text-following performance. In the second training stage, we incorporate all the training data and perform full-data tuning. This allows the model to further converge on all subtasks defined in this work.

After the second stage of full-data training, we observe that the generation quality is heavily influenced by the training data, particularly by low-quality training samples. To realign the generation quality with the generative prior of Flux, we design an image quality refinement training stage. Specifically, we utilize Flux to generate around $40K$ training samples. During training, we use the original images as references to guide the model in reconstructing itself. To prevent copy-paste effects, we drop $95\%$ tokens of reference images. After a shot-time optimization, the generation quality improved significantly, achieving alignment with the generation prior of Flux.

\begin{table}[t]
\caption{Quantitative evaluation of text-driven style transfer. The Text-sim$\uparrow$ is computed by the Cosine similarity between CLIP~\cite{clip} image embedding and CLIP text embedding.}
\vspace{-5pt}
\small
\centering
\setlength{\tabcolsep}{1.5pt}
\begin{tabular}{c | c c c c c c}
\toprule
 & StyleAlign & StyleShot & InstantStyle & DEADiff & CSGO & DreamO \\
\toprule
Style-sim $\uparrow$ & 0.7122 & 0.6922 & 0.6988 & 0.7269 & 0.7296 & \textbf{0.7340} \\
Text-sim $\uparrow$ & 0.2566 & 0.2693 & 0.2721 & 0.2656 & 0.2701 & \textbf{0.2750} \\
\toprule
\end{tabular}
\label{tb_cp_td}
\vspace{-5pt}
\end{table}

\section{Experiment}
\subsection{Implementation Details}
In this paper, we adopt Flux-1.0-dev as the base model. The rank of LoRA~\cite{lora} is set as 128, resulting in a total parameter increase of 478M. During training, we employ the Adam~\cite{adam} optimizer with a learning rate of 4e-5 and train on 8 NVIDIA A100 80G GPUs. The batch size is set as 8. The first training stage consists of 20K iterations, followed by 90K iterations in the second stage, and finally 3K iterations in the last training stage. In inference, we use Flux-Turbo~\cite{flux_turbo} for acceleration, enabling the generation of $1024\times 1024$ results within 10s. Unless specified, all results in this paper are based on the Turbo model. Some of the example inputs are processed using BEN2~\cite{ben2} to remove background.

\subsection{Qualitative Comparsion}
To validate the performance of DreamO, we conduct comparisons with recent state-of-the-art methods across multiple subtasks. The visual comparison is shown in Fig.~\ref{vis_cmp}. The first part presents a comparison between DreamO and SOTA identity-customization methods \textit{i.e.}, PhotoMaker~\cite{photomaker}, InstantID~\cite{instantid}, PuLID~\cite{pulid}. The results demonstrate that DreamO can inject identity information with high fidelity across various scenes, while offering impressive flexibility for customization.

The second part compares DreamO with recent subject-customization methods, including single-task frameworks (\textit{i.e.}, MS-Diffusion~\cite{ms_diffusion}) and unified generation frameworks (\textit{i.e.}, OmniGen~\cite{omnigen}, OminiControl~\cite{ominicontrol}). The results demonstrate that DreamO achieves higher subject fidelity and better text consistency in both single-subject and multi-subbject scenarios.

The third part shows comparison in virtual try-on, 
indicating that DreamO can effectively place clothing in scenes that align with the provided text with high fidelity to the reference images. Unlike methods like IMAGDressing, which produce high-fidelity clothing but lose text alignment, DreamO maintains both.

The last part presents the comparison between DreamO and recent style-customization methods, \textit{i.e.}, StyleShot~\cite{styleshot}, StyleAlign~\cite{stylealign}, InstantStyle~\cite{wang2024instantstyle}, DeaDiff~\cite{deadiff}, and CSGO~\cite{csgo}. It can be observed that DreamO has weaker content intrusion, better text alignment, and higher style fidelity in the generated results.

\begin{table}[t]
\captionsetup{type=table}
\caption{Quantitative evaluation of identity-driven customization.}
\vspace{-5pt}
\centering
\begin{tabular}{c | c c c c}
\toprule
 & PhotoMaker & InstantID & PuLID & DreamO \\
\toprule
Face-sim $\uparrow$ & 0.212 &  0.590 & 0.5829 & \textbf{0.607}\\
Text-sim $\uparrow$ & 0.2520 & 0.2294 & 0.2534 & \textbf{0.2570}\\
\toprule
\end{tabular}
\label{tb_cp_face}
\vspace{-5pt}
\end{table}

\begin{table}[t]
\caption{Quantitative evaluation of try-on.}
\vspace{-5pt}
\small
\centering
\setlength{\tabcolsep}{1.5pt}
\begin{tabular}{c | c c c c c}
\toprule
 & MagCloth & IMAGDressing & OmniGen & OminiControl & DreamO \\
\toprule
CLIP-sim $\uparrow$ & 0.5977 & \textbf{0.8405} &  0.7265 & 0.7065 & 0.7613\\
Text-sim $\uparrow$ & 30.17 & 17.74 & 27.83 & 28.79 & \textbf{30.47}\\
\toprule
\end{tabular}
\vspace{-5pt}
\label{tb_cp_cloth}
\end{table}

\begin{figure*}[t]
\centering
\begin{minipage}[t]{\linewidth}
\centering
\includegraphics[width=.9\columnwidth]{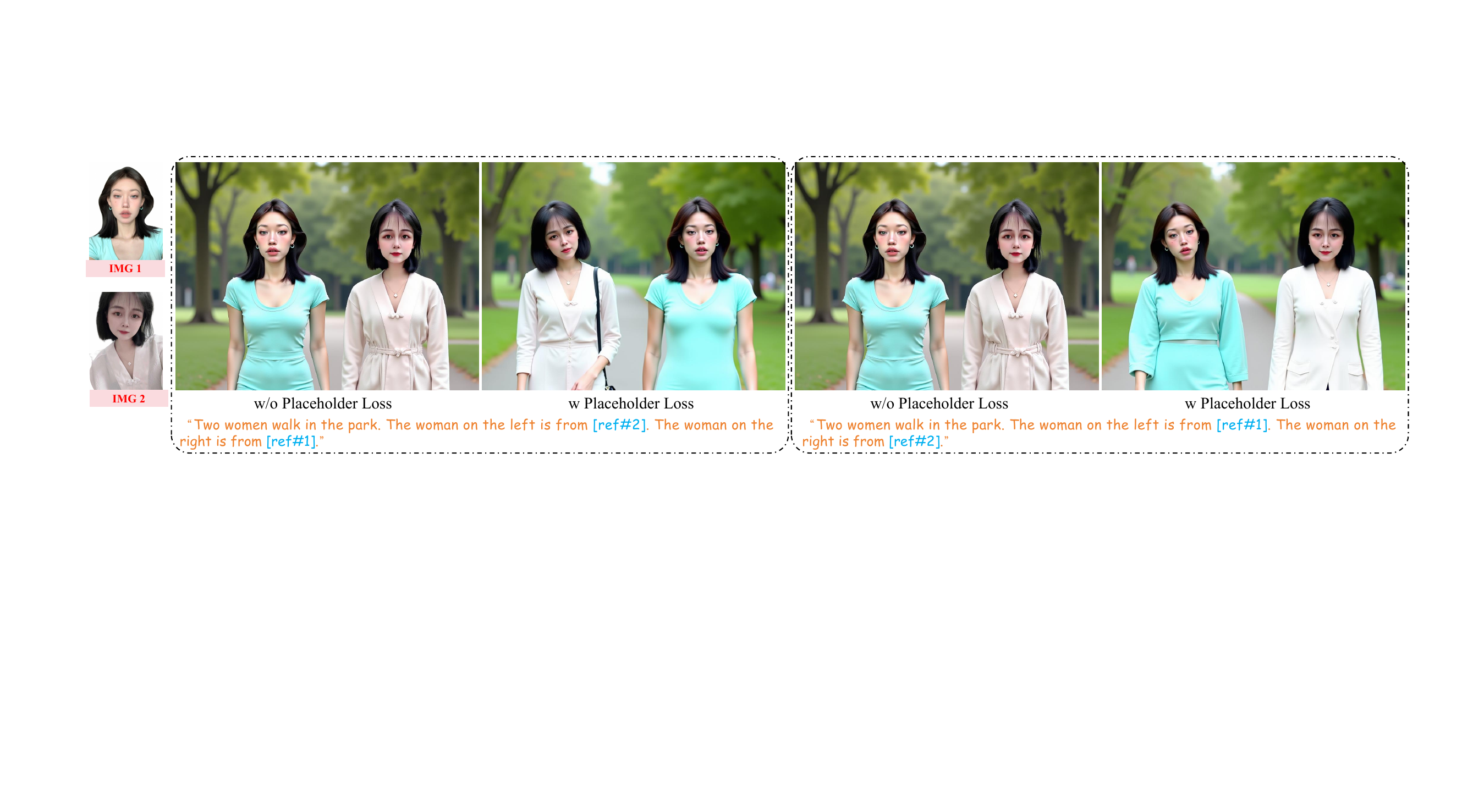}
\end{minipage}
\centering
\vspace{-15pt}
\caption{
The ablation study of the placeholder-to-image routing constraint.
}
\label{show_placeholder} 
\end{figure*}

\subsection{Quantitative Comparison}
In addition to qualitative comparison, we conduct quantitative comparisons for each task. We present the comparison of identity customization in Tab.~\ref{tb_cp_face}, which is evaluated on Unsplash-50~\cite{usp}. Here, we provide 9 prompts for each face. Following PuLID~\cite{pulid}, the Face-Sim represents the ID cosine similarity, with ID embeddings extracted by CurricularFace~\cite{cface}. We also compute the CLIP cosine similarity between the generated result and the prompt to measure the text-following ability of different methods. As can be seen, our DreamO shows better face similarity and text-based customization ability.

Tab.~\ref{tb_cp_subject} presents the quantitative comparison in single- and multi-subject customization. We use DreamBench~\cite{dreambooth} as the testset of single-subject customization. For multi-subject customization, we randomly select 20 pairs from DreamBench and provide 25 prompts for each. During testing, we generate four images with different seeds for each test sample. Here, we calculate the CLIP cosine similarity and Dino~\cite{dino} cosine similarity between the generated result and the reference images as a measure of subject consistency. To improve accuracy, we remove the background of the generated results and then calculate similarity. Additionally, we employ CLIP cosine similarity between the text description and the generated result as a measure of content alignment. Tab.~\ref{tb_cp_subject} shows that our method outperforms others in subject consistency while demonstrating strong text-following ability.

The quantitative comparison of the try-on is presented in Tab.~\ref{tb_cp_cloth}. We select 300 reference garments from VITON-HD~\cite{viton} encompassing various styles and colors as the test data. During testing, we provide 10 prompts for each cloth. The CLIP consine similarity between the generated result and reference cloth is employed to measure the try-on accuracy. Here, we only crop the cloth from the result to compute the CLIP-sim. The CLIP cosine similarity between the generated results and the prompt is employed to measure the text-following ability of different methods. The result in Tab.~\ref{tb_cp_cloth} shows the attractive performance of our DreamO. Although IMAGDressing~\cite{imdress} has higher CLIP-sim, it can only generate images with a white background (\textit{i.e.}, Fig.~\ref{vis_cmp} ) with little text-following ability.

For style customization, we construct an evaluation dataset, containing $249$ style images and $24$ prompt. Each method generates $249\times 24$ style customization results. We use the pre-trained CSD~\cite{csd} to extract the style feature of generated results and reference images, and calculate the cosine similarity between them as a measure of style consistency. Furthermore, we compute the CLIP~\cite{clip} similarity between stylized results and text descriptions as the metric of content consistency. The results are shown in Tab.~\ref{tb_cp_td}, which demonstrate that our method has better performance in style consistency and content consistency.

\begin{figure}[t]
\centering
\begin{minipage}[t]{\linewidth}
\centering
\includegraphics[width=1\columnwidth]{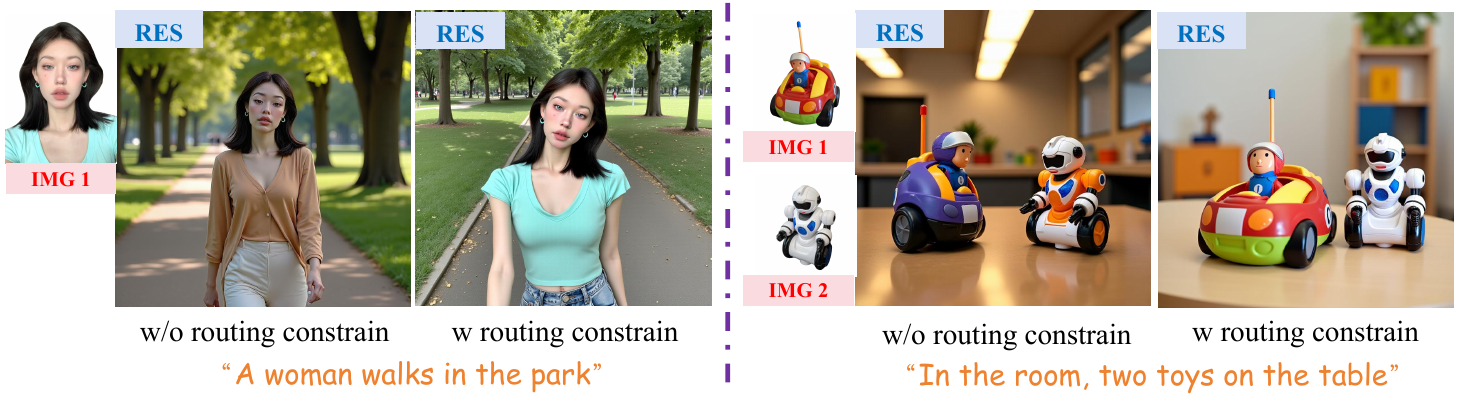}
\end{minipage}
\centering
\vspace{-10pt}
\caption{
The ablation study of routing constraint in our proposed DreamO.
}
\label{abs_route} 
\end{figure}

\begin{figure}[t]
\centering
\begin{minipage}[h]{\linewidth}
\centering
\includegraphics[width=1\columnwidth]{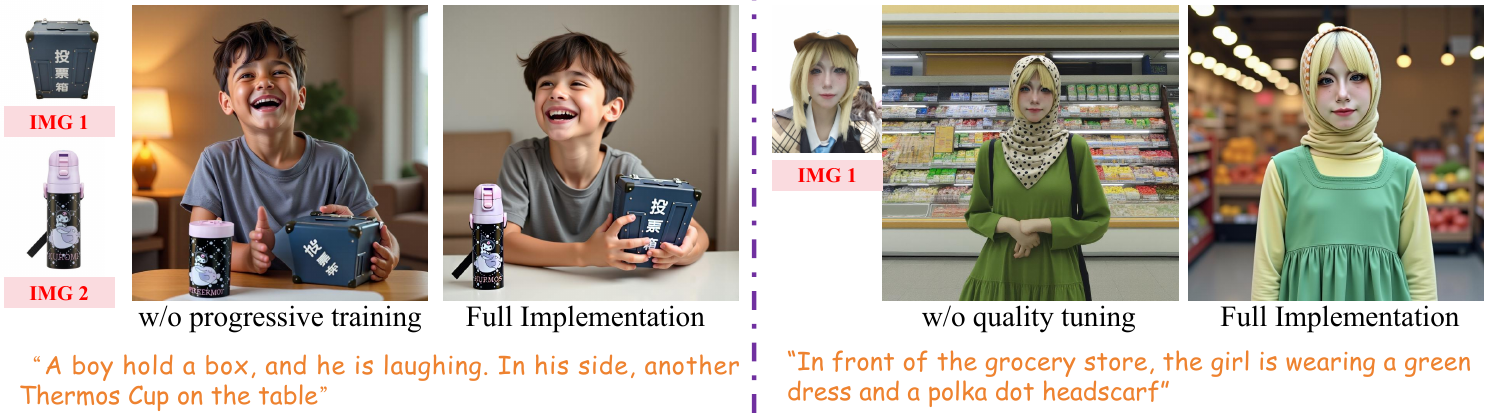}
\end{minipage}
\centering
\vspace{-10pt}
\caption{
The ablation study of progressive training in our proposed DreamO.
}
\label{abs_pt} 
\end{figure}

\subsection{User Study}
\begin{figure}[t]
\centering
\begin{minipage}[h]{\linewidth}
\centering
\includegraphics[width=1\columnwidth]{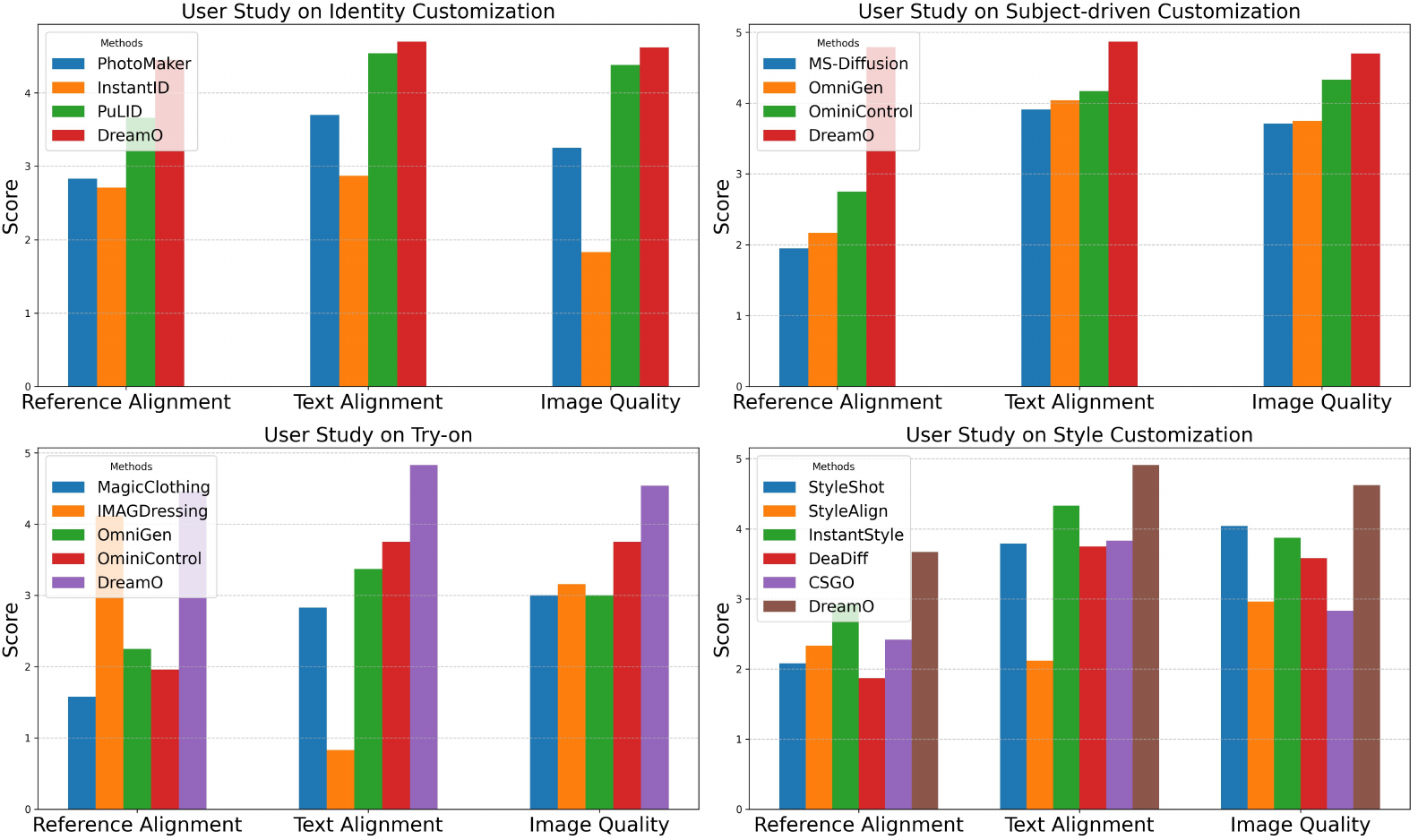}
\end{minipage}
\centering
\caption{
The user study of different methods.
}
\vspace{-10pt}
\label{fig_user} 
\end{figure}

In addition to the automatic evaluation metrics, we also conduct a user study for manual evaluation of different methods. Specifically, for each task (\textit{i.e.}, style, object, identity, and try-on customization), we assigned 6 test samples and invited 20 volunteers to rate on three aspects: text alignment, reference alignment, and image quality. The scoring range was set from 0 to 5, where a higher score indicates greater satisfaction. Fig.~\ref{fig_user} shows that our DreamO achieves better performance in these three evaluation aspects.

\subsection{Ablation Study}

\begin{table}[t]
\captionsetup{type=table}
\caption{Ablation study of different model settings in multi-subject-driven customization.}
\vspace{-10pt}
\centering
\begin{tabular}{c | c c c c}
\toprule
 & w/o CE & w/o RC & w/o PT & DreamO \\
\toprule
CLIP-sim $\uparrow$ & 0.7697 & 0.7448 & 0.7349 & \textbf{0.7775} \\
DINO-sim $\uparrow$ & 0.6097 & 0.5540 & 0.5381 & \textbf{0.6253}\\
Text-sim $\uparrow$ & 31.26 & 28.42 & 28.31 & \textbf{31.46}\\
\toprule
\end{tabular}
\vspace{-10pt}
\label{supp_abs}
\end{table}

\textbf{Routing constraint}. In this paper, we introduce a routing constraint into DiT training to enhance generation fidelity and facilitate the decoupling of multi-condition control. To evaluate its effectiveness, we ablate the routing constraint during training, with results shown in Fig.~\ref{abs_route}. In single-condition generation, its removal leads to degraded reference fidelity, \textit{e.g.}, the clothing color becomes inconsistent with the reference. In multi-condition settings, it causes condition coupling, \textit{e.g.}, features of the two toys are crossed. These results confirm that the routing constraint improves the fidelity and disentanglement of different conditions.

\noindent \textbf{Progressive training}. To enable the model to better converge on all sub-tasks under complex data distributions and to rectify the impact of training data distribution on generation quality, we design a progressive training strategy. The effectiveness of this strategy is demonstrated in Fig.~\ref{abs_pt}. One can see that directly training the model on all datasets leads to suboptimal convergence, particularly in complex tasks such as multi-subject consistency. Warming up on a smaller and easier-to-learn dataset (\textit{e.g.}, Subject200K~\cite{ominicontrol}) before joint training improves convergence, but the generation quality is easily influenced by the training data distribution, deviating from the generation priors of Flux. By introducing an image quality tuning stage, the model can produce higher-quality generation results.

\noindent \textbf{Placeholder-to-image routing constraint}. As demonstrated in Eq.~\ref{loss_placeholder}, this paper designs a placeholder-to-image routing constraint to build the routing relationship between placeholders and specific images. Fig.~\ref{show_placeholder} shows the effect of this loss term. It can be seen that without this loss, placeholders struggle to precisely control their corresponding images. After applying this loss, placeholders can bind to specific reference objects, enabling individual control over particular objects during multi-subject customization.

\noindent \textbf{Quantitative results}. In addition to the visual comparison, we show the quantitative results of the ablation study, as shown in Tab.~\ref{supp_abs}. The experiment is conducted on multi-subject-driven customization. It can be seen that not using the routing constraint (RC) and progressive training strategy (PT) significantly impacts performance, leading to a decrease in the reference consistency and text following. We also study the role of condition embedding (CE), and its absence results in a decline in the reference consistency.

\section{Conclusion}

In this study, we introduce DreamO, a unified framework designed for generalized image customization across diverse condition types (e.g., identity, style, subject, and try-on) within a single pre-trained DiT architecture. To facilitate this, we construct a large-scale training dataset. By embedding all condition types into the DiT input sequence and incorporating a feature routing constraint, DreamO achieves high-fidelity consistency while effectively disentangling heterogeneous control signals. 
In addition, we design a progressive training strategy that enables the model to incrementally acquire diverse control capabilities under complex data distributions, while maintaining the image quality inherent to the base model.
Comprehensive experiments demonstrate that DreamO excels in performing a wide range of image customization tasks with high-quality results.

\bibliographystyle{ACM-Reference-Format}
\bibliography{sample-bibliography}

\clearpage
\newpage

\begin{figure*}[t]
\centering
\begin{minipage}[t]{\linewidth}
\centering
\includegraphics[width=1\columnwidth]{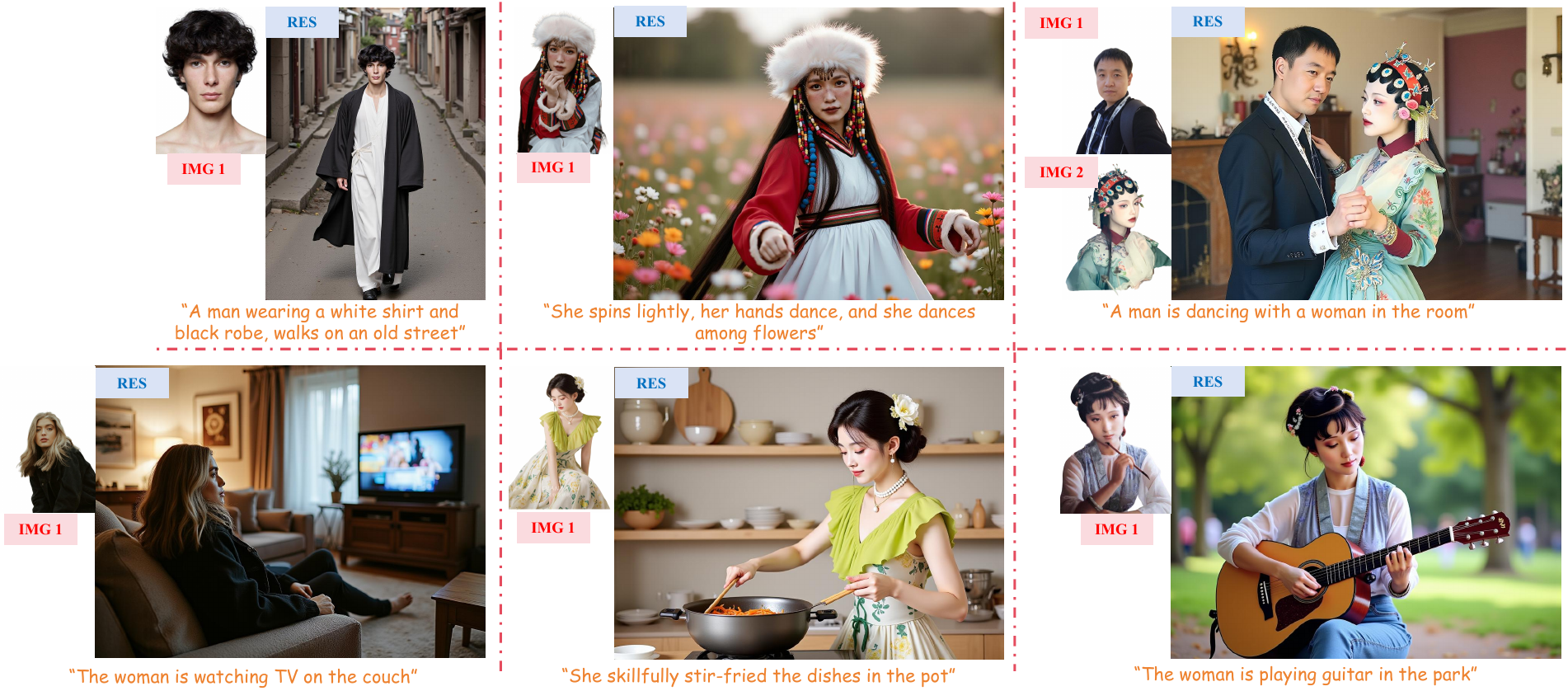}
\end{minipage}
\centering
\caption{
The capability of our proposed DreamO in identity-driven image customization.
}
\label{show_1} 
\end{figure*}

\begin{figure*}[t]
\centering
\begin{minipage}[t]{\linewidth}
\centering
\includegraphics[width=1\columnwidth]{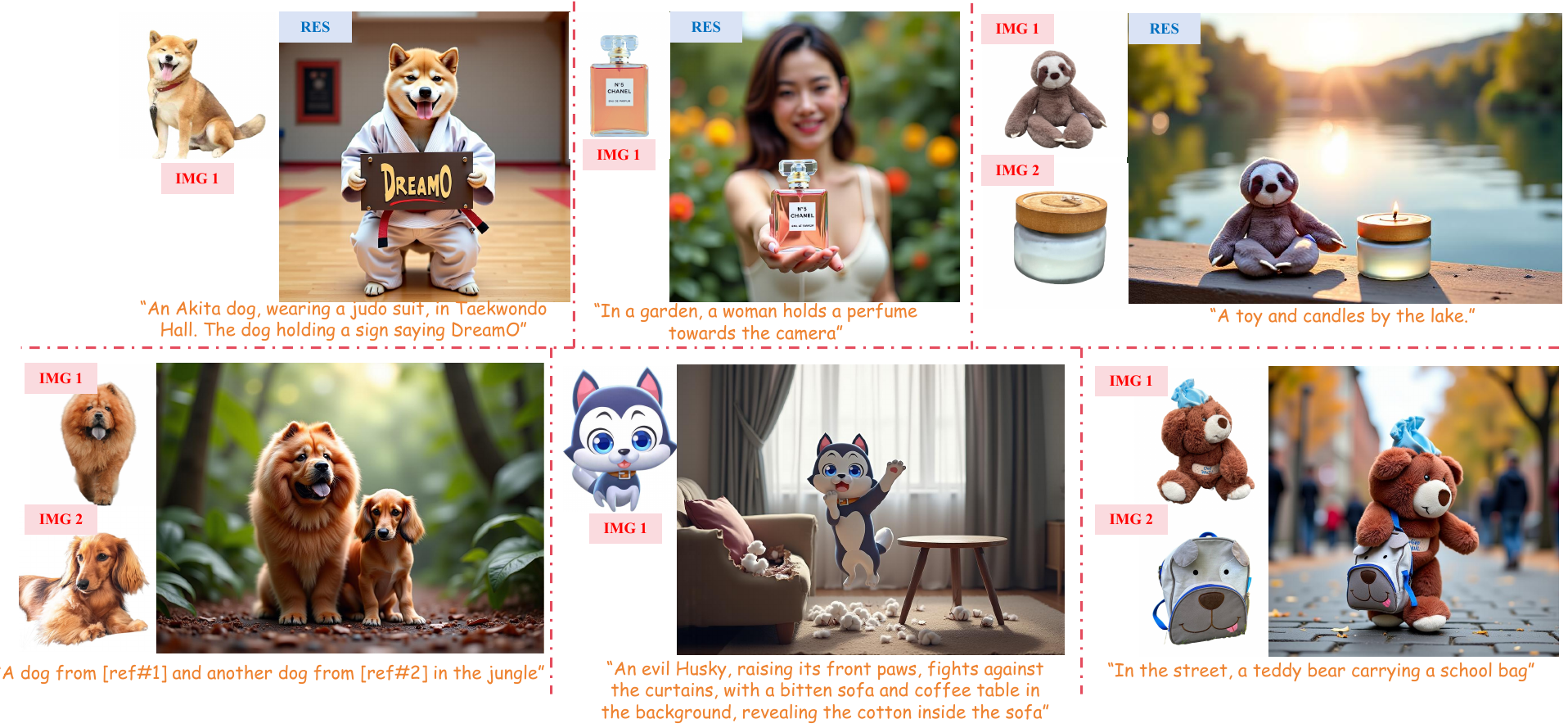}
\end{minipage}
\centering
\caption{
The capability of our proposed DreamO in subject-driven image customization.
}
\label{show_2} 
\end{figure*}

\begin{figure*}[t]
\centering
\begin{minipage}[t]{\linewidth}
\centering
\includegraphics[width=1\columnwidth]{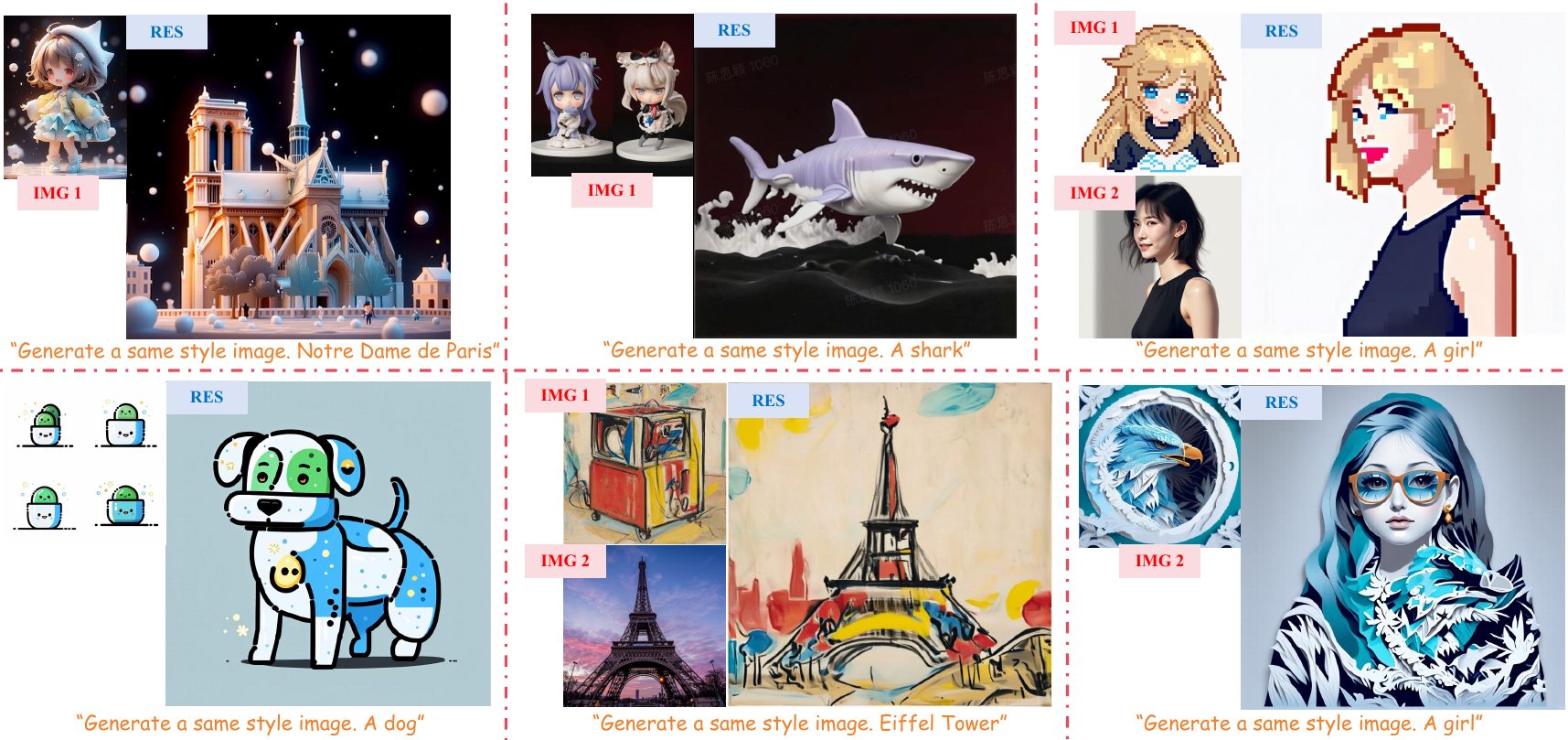}
\end{minipage}
\centering
\caption{
The capability of our proposed DreamO in style-driven image customization.
}
\label{show_4} 
\end{figure*}

\end{document}